\documentclass[sigplan,nonacm]{acmart}

\usepackage{amsmath}
\usepackage{amsfonts}
\usepackage{booktabs}
\usepackage{graphicx}
\usepackage{multirow}
\usepackage{multicol}
\usepackage{tablefootnote}
\usepackage{float}
\usepackage{url}
\usepackage{makecell}
\usepackage{hyperref}
\usepackage[capitalize,noabbrev]{cleveref}
\usepackage{subcaption}

\usepackage{xspace}

\AtBeginDocument{%
  }

\setcopyright{acmlicensed}
\copyrightyear{2018}
\acmYear{2018}
\acmDOI{XXXXXXX.XXXXXXX}
\acmConference[BIOKDD 2025]{24th International Workshop on Data Mining in Bioinformatics}{June 03--05,
  2018}{Woodstock, NY}
\acmISBN{978-1-4503-XXXX-X/2018/06}

\begin{document}


\title{Binary Latent Protein Fitness Landscapes for Quantum Annealing Optimization}

\author{Truong-Son Hy}
\affiliation{
  \institution{University of Alabama at Birmingham}
  \city{Birmingham}
  \state{Alabama}
  \country{USA}
}
\email{thy@uab.edu}

\newcommand{\modelname}{\textsc{Q-BioLat}\xspace}

\renewcommand{\shortauthors}{Truong-Son Hy}

\begin{abstract}
We propose \modelname, a framework for modeling and optimizing protein fitness landscapes in binary latent spaces. Starting from protein sequences, we leverage pretrained protein language models to obtain continuous embeddings, which are then transformed into compact binary latent representations. In this space, protein fitness is approximated using a quadratic unconstrained binary optimization (QUBO) model, enabling efficient combinatorial search via classical heuristics such as simulated annealing and genetic algorithms.

On the ProteinGym benchmark, we demonstrate that \modelname captures meaningful structure in protein fitness landscapes and enables the identification of high-fitness variants. Despite using a simple binarization scheme, our method consistently retrieves sequences whose nearest neighbors lie within the top fraction of the training fitness distribution, particularly under the strongest configurations. We further show that different optimization strategies exhibit distinct behaviors, with evolutionary search performing better in higher-dimensional latent spaces and local search remaining competitive in preserving realistic sequences.

Beyond its empirical performance, \modelname provides a natural bridge between protein representation learning and combinatorial optimization. By formulating protein fitness as a QUBO problem, our framework is directly compatible with emerging quantum annealing hardware, opening new directions for quantum-assisted protein engineering.

Our implementation is publicly available at: \url{https://github.com/HySonLab/Q-BIOLAT}.

\end{abstract}



\maketitle

\section{Introduction} \label{sec:Introduction}

\begin{figure*}
    \centering
    \includegraphics[width=0.9\linewidth]{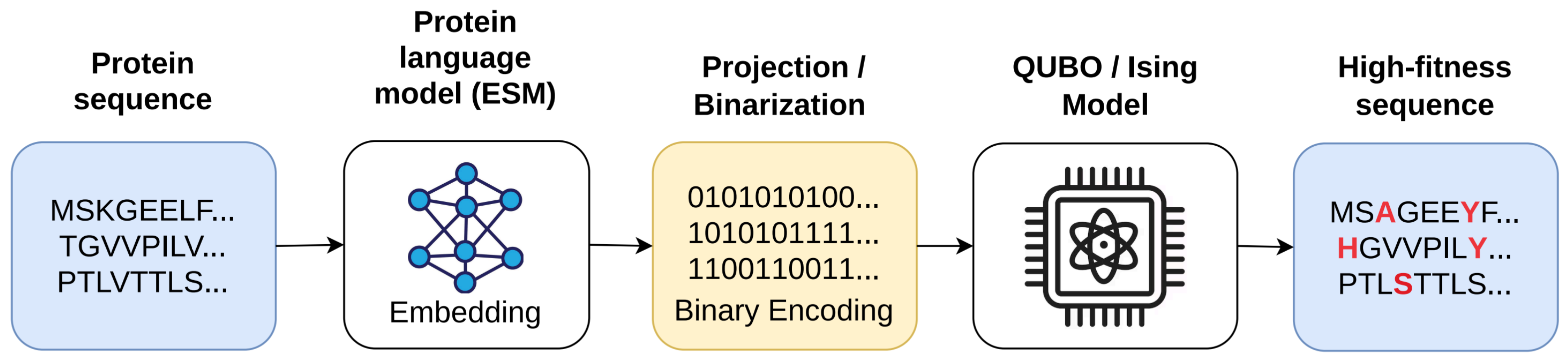}
    \caption{Overview of the \modelname framework. Protein sequences are first encoded using a pretrained protein language model (ESM) to obtain continuous embeddings. These embeddings are transformed into binary latent representations through projection and binarization, enabling protein fitness to be modeled as a quadratic unconstrained binary optimization (QUBO) problem. The resulting latent fitness landscape can be explored using combinatorial optimization methods and is directly compatible with quantum annealing hardware. Optimized latent codes are mapped back to high-fitness protein sequences.}
    \label{fig:Main_Figure}
\end{figure*}

Understanding and optimizing protein fitness landscapes is a central challenge in computational biology, with applications spanning enzyme engineering, drug discovery, and synthetic biology. Protein fitness landscapes are typically high-dimensional, rugged, and shaped by complex interactions between residues, making efficient exploration of sequence space difficult \cite{romero2009exploring, CHEN2023706, ProteinGym}. Recent advances in protein language models (PLMs), such as ESM-2 \cite{ESM-2} and ESM-3 \cite{ESM-3}, have enabled powerful representations of protein sequences by learning from large-scale unlabeled data. These representations have significantly improved performance on downstream tasks, including mutation effect prediction \cite{10.1093/bioinformatics/btae621, liu2025plm} and protein design \cite{watson2023novo, LatentDE, DiscreteDE}.

Despite these advances, most existing approaches rely on continuous neural predictors and gradient-based \cite{LatentDE} or sampling-based optimization strategies \cite{DiscreteDE}, which are not naturally suited for discrete combinatorial search. In particular, protein sequences are inherently discrete objects, and many biologically relevant optimization problems can be more naturally formulated in discrete spaces. Classical approaches such as evolutionary algorithms \cite{DiscreteDE, OZAWA2024101758} and Bayesian optimization \cite{atkinson2025protein, 10.1093/bib/bbac570} have been widely applied to protein engineering, but they often struggle with scalability or require large numbers of evaluations.

In this work, we introduce \modelname, a framework that maps protein sequences into binary latent spaces where fitness landscapes can be explicitly modeled and optimized. Starting from pretrained protein language model embeddings, we construct compact binary representations that enable the formulation of protein fitness as a quadratic unconstrained binary optimization (QUBO) \cite{kochenberger2014unconstrained, IsingFormulation, kim2025quantum} problem. This formulation allows us to leverage a wide range of combinatorial optimization techniques, including simulated annealing and genetic algorithms, for efficient exploration of protein fitness landscapes.

We evaluate \modelname on real protein fitness data from the ProteinGym benchmark \cite{ProteinGym} and show that even simple binary latent representations capture meaningful structure in protein fitness landscapes. Our results demonstrate that combinatorial optimization in latent space can consistently identify high-fitness variants, with retrieved sequences lying in the top fraction of the training fitness distribution. Furthermore, we observe that different optimization strategies exhibit distinct behaviors depending on the dimensionality of the latent space, highlighting the importance of representation design for effective search.

Beyond its empirical performance, \modelname provides a natural bridge between protein representation learning and combinatorial optimization. By expressing protein fitness landscapes as QUBO problems, our framework is directly compatible with emerging quantum annealing hardware, offering a potential pathway toward quantum-assisted protein engineering. While our experiments focus on classical optimization methods, the proposed formulation opens new directions for integrating quantum optimization techniques with modern protein language models.

In summary, this work demonstrates that binary latent representations can transform protein fitness prediction into a structured combinatorial optimization problem, enabling new opportunities for efficient search and future integration with quantum computing paradigms.

\section{Related Work} \label{sec:related_works}

\textbf{Protein language models and sequence representations.}
Recent advances in protein language models (PLMs) have significantly improved the representation of biological sequences by leveraging large-scale unlabeled protein data. Models such as ESM-2 \cite{ESM-2} and its successor ESM-3 \cite{ESM-3} learn contextual embeddings that capture structural and functional properties of proteins directly from sequence data. More recently, multimodal protein language models have been developed to integrate sequence information with additional modalities such as structure, function, and evolutionary context, further enhancing representation quality and downstream performance \cite{10.1093/biomethods/bpae043, Khang_Ngo_2024}. These models have demonstrated strong performance across a wide range of tasks, including structure prediction, mutation effect prediction, and protein design \cite{doi:10.1073/pnas.2016239118, lin2023evolutionary}. In this work, we build upon PLM embeddings as a foundation, but transform them into binary latent representations to enable discrete optimization. \\

\noindent
\textbf{Protein fitness prediction and engineering.}
Predicting protein fitness landscapes is a fundamental problem in computational biology, with applications in enzyme optimization, therapeutic design, and synthetic biology. Recent approaches combine machine learning models with experimental data, particularly from deep mutational scanning (DMS) experiments, to predict the effects of mutations \cite{yang2019machine, biswas2021low}. Large-scale benchmarks such as ProteinGym have been introduced to systematically evaluate fitness prediction models across diverse proteins and mutation regimes \cite{ProteinGym}. While these methods achieve strong predictive performance, they typically rely on continuous representations and do not directly address the combinatorial nature of sequence optimization. Furthermore, these approaches primarily focus on prediction rather than explicitly enabling structured optimization over discrete sequence spaces. \\

\noindent
\textbf{Optimization methods for protein design.}
A variety of optimization strategies have been applied to protein engineering, including evolutionary algorithms \cite{LatentDE, DiscreteDE, OZAWA2024101758, 10.1093/bib/bbac570}, Bayesian optimization \cite{10.1093/bib/bbac570, atkinson2025protein}, and reinforcement learning \cite{wang2025reinforcement, sun2025accelerating}. Evolutionary methods such as genetic algorithms are widely used due to their ability to explore large discrete search spaces, while Bayesian optimization offers sample-efficient search under expensive evaluation settings. Classical optimization techniques such as simulated annealing have also been applied to combinatorial problems with rugged energy landscapes \cite{kirkpatrick1983optimization, 10.7551/mitpress/1090.001.0001}. However, many of these approaches operate directly in sequence space or continuous embedding spaces, rather than explicitly modeling protein fitness as a structured combinatorial optimization problem. In contrast, our approach explicitly models protein fitness as a structured combinatorial optimization problem in a learned binary latent space. \\

\noindent
\textbf{QUBO formulations and quantum annealing.}
Quadratic unconstrained binary optimization (QUBO) \cite{kochenberger2014unconstrained, IsingFormulation, kim2025quantum} provides a unified framework for expressing a wide range of combinatorial problems in terms of binary variables and pairwise interactions. QUBO formulations are closely related to the Ising model and form the basis of many optimization techniques, including quantum annealing \cite{IsingFormulation}. Quantum annealing hardware, such as that developed by D-Wave Systems, has demonstrated the ability to solve certain classes of optimization problems by exploiting quantum effects \cite{Johnson2011QuantumAW}. While applications of quantum optimization in computational biology remain limited, recent work has begun to explore its potential for tasks such as molecular design and protein folding. In this work, we formulate protein fitness landscapes as QUBO problems in a binary latent space, enabling direct compatibility with both classical combinatorial solvers and emerging quantum annealing hardware, and providing a unified perspective on protein design as discrete energy landscape optimization. \\

\noindent
\textbf{Latent space optimization and discrete representations.}
Optimizing in learned latent spaces has become a common strategy in machine learning, particularly for generative modeling and design tasks \cite{hottung2021learning, 10.1145/3520304.3533993}. Continuous latent spaces have been widely used for molecule and protein generation \cite{doi:10.1021/acsomega.2c03264, MGVAE, LatentDE}, but they often require gradient-based optimization or sampling methods that do not naturally handle discrete constraints. More recently, discrete and binary latent representations have been explored for enabling combinatorial search and efficient optimization \cite{jiang2023efficient, abe2026effectiveness}. Our work builds on this idea by constructing binary latent representations derived from protein language models and explicitly modeling the resulting fitness landscape as a QUBO problem, bridging representation learning and combinatorial optimization.
\section{Method}

\subsection{Overview}

We propose \modelname, a framework for modeling and optimizing protein fitness landscapes in a binary latent space. Given a protein sequence, \modelname first computes a pretrained protein language model embedding, then transforms this continuous representation into a compact binary latent code. In this binary space, the fitness landscape is approximated by a quadratic unconstrained binary optimization (QUBO) surrogate, which models both unary and pairwise effects among latent variables. The learned surrogate is then optimized using combinatorial search methods such as simulated annealing and genetic algorithms. Because the final objective is expressed in QUBO form, the framework is naturally compatible with both classical combinatorial solvers and quantum annealing hardware.

Formally, the overall pipeline is
\[
s \;\rightarrow\; e(s) \in \mathbb{R}^{d}
\;\rightarrow\; z(s) \in \mathbb{R}^{m}
\;\rightarrow\; x(s) \in \{0,1\}^{m}
\;\rightarrow\; \hat{f}(x),
\]
where $s$ is a protein sequence, $e(s)$ is the protein language model embedding, $z(s)$ is a reduced continuous latent representation, $x(s)$ is the binarized latent code, and $\hat{f}(x)$ is the QUBO surrogate of protein fitness.

\subsection{Problem Setup}

Let $\mathcal{D}=\{(s_i,y_i)\}_{i=1}^{N}$ denote a protein fitness dataset, where $s_i$ is a protein sequence and $y_i \in \mathbb{R}$ is its experimentally measured fitness. Our goal is to learn a surrogate model that maps each sequence to a compact binary latent representation and predicts its fitness, while also enabling efficient combinatorial optimization over the latent space.

Unlike conventional approaches that optimize directly in sequence space or in continuous embedding space, we aim to transform protein fitness prediction into a discrete optimization problem. This is motivated by two observations: first, protein sequences are inherently discrete objects; second, binary latent spaces admit efficient combinatorial search and can be directly mapped to QUBO and Ising formulations.

\subsection{Protein Language Model Embeddings}

For each protein sequence $s_i$, we obtain a continuous embedding using a pretrained protein language model. In our experiments, we use ESM-based sequence embeddings \cite{ESM-2, ESM-3} due to their strong empirical performance and broad adoption in protein representation learning. Given a sequence $s_i$ of length $L_i$, the language model produces contextualized residue-level representations
\[
H_i \in \mathbb{R}^{L_i \times d},
\]
where $d$ is the hidden dimension of the model. To obtain a fixed-length sequence representation, we apply mean pooling across residues:
\[
e_i = \frac{1}{L_i} \sum_{j=1}^{L_i} H_i^{(j)} \in \mathbb{R}^{d}.
\]
This yields one dense embedding vector per sequence.

The role of the protein language model in \modelname is not to directly predict fitness, but rather to provide a biologically informed continuous representation that can later be compressed into a discrete latent code suitable for combinatorial optimization.

\subsection{Continuous-to-Binary Latent Mapping}

\paragraph{Dimensionality reduction.}
The dense embedding $e_i \in \mathbb{R}^{d}$ is first mapped into a lower-dimensional latent space of dimension $m \ll d$. We consider two strategies in our experiments:
\begin{itemize}
    \item \textbf{Random projection:} 
    \[
    z_i = W e_i, \qquad W \in \mathbb{R}^{m \times d},
    \]
    where $W$ is sampled from a random Gaussian distribution with variance scaled by $1/d$.
    
    \item \textbf{Principal component analysis (PCA):} the top $m$ principal directions are fitted on the embedding matrix and used to obtain
    \[
    z_i = \mathrm{PCA}(e_i) \in \mathbb{R}^{m}.
    \]
\end{itemize}
The purpose of this step is twofold: to compress the representation and to control the size of the downstream QUBO model, whose parameter count scales quadratically with the latent dimension.

\paragraph{Binarization.}
The reduced continuous latent vector $z_i \in \mathbb{R}^{m}$ is then transformed into a binary code
\[
x_i \in \{0,1\}^{m}.
\]
In our main experiments, we use median-threshold binarization:
\[
x_{ik} = \mathbb{I}\big(z_{ik} > \tau_k\big),
\]
where $\tau_k$ is the median of the $k$-th latent component computed over the training set, and $\mathbb{I}(\cdot)$ is the indicator function. This choice yields approximately balanced binary activations across dimensions and avoids degenerate codes with highly imbalanced bit frequencies.

The resulting binary representation defines a discrete latent space in which each sequence corresponds to a vertex of the Boolean hypercube.

\subsection{QUBO Surrogate for Protein Fitness}

We model protein fitness in binary latent space with a QUBO surrogate. Given a binary latent code $x \in \{0,1\}^{m}$, the predicted fitness is:
\[
\hat{f}(x) = \sum_{k=1}^{m} h_k x_k + \sum_{1 \leq k < \ell \leq m} J_{k\ell} x_k x_\ell,
\]
where $h_k \in \mathbb{R}$ captures the unary contribution of latent bit $k$, and $J_{k\ell} \in \mathbb{R}$ captures the pairwise interaction between bits $k$ and $\ell$. Equivalently, the model can be written in matrix form as:
\[
\hat{f}(x) = h^\top x + \frac{1}{2}x^\top J x,
\]
where $J \in \mathbb{R}^{m \times m}$ is the symmetric Hamiltonian (i.e. pairwise interaction matrix) with zero diagonal, and $h \in \mathbb{R}^m$ corresponds to the bias term. This representation connects the latent fitness model to the classical QUBO formulation.

\paragraph{Feature construction.}
For each binary code $x_i$, we construct a feature vector consisting of:
\begin{itemize}
    \item all linear terms $\{x_{ik}\}_{k=1}^{m}$,
    \item all pairwise interaction terms $\{x_{ik}x_{i\ell}\}_{1 \leq k < \ell \leq m}$.
\end{itemize}
The total number of features is therefore
\[
m + \frac{m(m-1)}{2}.
\]

\paragraph{Parameter estimation.}
We fit the QUBO surrogate using ridge regression. Let $\Phi(X)$ denote the design matrix formed by the linear and pairwise features of the training binary codes, and let $y$ denote the corresponding fitness vector. We solve
\[
w^\star = \arg\min_{w} \|\Phi(X)w - y\|_2^2 + \lambda \|w\|_2^2,
\]
where $\lambda > 0$ is an $\ell_2$ regularization coefficient. The learned parameter vector $w^\star$ is then unpacked into the unary coefficients $h$ and pairwise coefficients $J$.

This surrogate is attractive for three reasons. First, it is interpretable, since each term corresponds to a unary or pairwise latent effect. Second, it is computationally efficient to fit and evaluate for moderate latent dimensions. Third, it directly yields a QUBO objective suitable for classical and quantum combinatorial optimization.

\subsection{Latent Space Optimization}

Once the QUBO surrogate has been fitted, we seek binary latent codes that maximize the predicted fitness:
\[
x^\star = \arg\max_{x \in \{0,1\}^{m}} \hat{f}(x).
\]
Because this is a discrete combinatorial problem, we employ search strategies that operate directly in the binary latent space.

\paragraph{Simulated annealing \cite{kirkpatrick1983optimization}.}
Simulated annealing starts from an initial latent code and iteratively proposes single-bit flips. Moves that improve the objective are accepted, while worse moves may be accepted with a temperature-dependent probability, enabling escape from local optima. The temperature is gradually reduced during the search.

\paragraph{Genetic algorithm \cite{10.7551/mitpress/1090.001.0001}.}
The genetic algorithm maintains a population of binary latent codes and evolves them through selection, crossover, and mutation. This allows broader exploration of the latent space and is particularly useful in higher-dimensional settings where multiple promising regions may exist.

\paragraph{Baselines.}
To contextualize performance, we also compare against greedy hill climbing, random search, and a lightweight latent Bayesian-style search baseline. Greedy hill climbing iteratively flips the single bit that most improves the objective until convergence. Random search samples binary codes uniformly and retains the best candidate. The latent Bayesian-style method uses a kernel-based uncertainty heuristic over binary latent codes.

\subsection{Retrieval-Based Decoding of Optimized Latents}

A practical challenge in binary latent optimization is how to map an optimized latent code back to a protein sequence. In this work, we adopt a retrieval-based decoding strategy rather than training a generative decoder. Given an optimized latent code $x^\star$, we retrieve its nearest observed training sequences under Hamming distance:
\[
d_H(x^\star, x_i) = \sum_{k=1}^{m} |x_k^\star - x_{ik}|.
\]
The retrieved nearest neighbors serve as real sequences associated with the optimized region of latent space. This strategy provides a conservative and interpretable way to evaluate whether the optimization procedure steers search toward high-fitness regions of the observed sequence manifold.

We emphasize that retrieval-based decoding is intentionally simple and avoids introducing an additional generative model. Its role in this paper is to provide a first practical proxy for mapping optimized latent solutions back to real protein sequences.

\subsection{Quantum Annealing Compatibility}

An important property of \modelname is that its optimization objective is already expressed in QUBO form. This means the learned latent fitness landscape can be optimized not only with classical combinatorial solvers, but also with quantum annealing hardware or other Ising/QUBO solvers without changing the model formulation. In this sense, \modelname provides a quantum-compatible interface between protein representation learning and discrete optimization.

In the present work, we focus on classical solvers to establish the feasibility of the framework and to analyze the optimization behavior in binary latent space. Nevertheless, the QUBO formulation opens a clear pathway toward future quantum-assisted protein engineering, where learned latent fitness landscapes could be deployed on quantum annealers or related specialized optimization hardware.

\subsection{Computational Complexity}

Let $m$ denote the binary latent dimension. The QUBO surrogate uses
\[
m + \frac{m(m-1)}{2} = \mathcal{O}(m^2)
\]
features. Therefore, increasing latent dimensionality increases expressive power but also increases both the number of surrogate parameters and the difficulty of combinatorial search. This trade-off motivates our experimental study over multiple latent dimensions.

By separating the dense protein language model embedding stage from the binary optimization stage, \modelname keeps the most expensive neural computation fixed while enabling fast repeated optimization in the compact binary latent space. This makes the framework computationally attractive for studying representation--optimization trade-offs.
\section{Experiments} \label{sec:experiments}

\subsection{Dataset}

We evaluate \modelname on protein fitness prediction tasks using data from the ProteinGym benchmark \cite{ProteinGym}, which aggregates large-scale deep mutational scanning (DMS) experiments across diverse proteins. In our experiments, we focus on the GFP (green fluorescent protein) dataset, which contains experimentally measured fitness values for a large number of sequence variants.

To study the effect of dataset size, we construct subsets of size $\{1000, 2000, 5000, 10000\}$ by random sampling with a fixed seed. Each dataset is split into training and test sets using an 80/20 split. All reported results are averaged over multiple random seeds (i.e. 5) unless otherwise specified.

\subsection{Experimental Setup}

\paragraph{Protein representations.}
We obtain protein embeddings using a pretrained protein language model (ESM-2) \cite{ESM-2}. For each sequence, we compute a fixed-length representation via mean pooling over residue-level embeddings.

\paragraph{Latent representation.}
We project embeddings into a lower-dimensional latent space with dimension $m \in \{8, 16, 32, 64\}$ using either random projection or principal component analysis (PCA). The resulting continuous vectors are binarized using median thresholding to produce binary latent codes.

\paragraph{QUBO surrogate.}
We fit a quadratic surrogate model over binary latent codes using ridge regression. The surrogate includes both linear terms and pairwise interactions, corresponding to a QUBO formulation.

\paragraph{Optimization methods.}
We compare the following optimization strategies:

\begin{itemize}
    \item Simulated annealing (SA)
    \item Genetic algorithm (GA)
    \item Greedy hill climbing
    \item Random search
    \item Latent Bayesian-style search (BO)
\end{itemize}

\paragraph{Evaluation metrics.}
We evaluate performance using:

\begin{itemize}
    \item \textbf{Spearman correlation}: measures ranking accuracy of the surrogate.
    \item \textbf{Surrogate improvement}: increase in predicted fitness relative to the starting point.
    \item \textbf{Nearest-neighbor true fitness}: fitness of the closest observed sequence to the optimized latent code.
    \item \textbf{Nearest-neighbor percentile}: percentile rank of the retrieved sequence within the training set.
\end{itemize}

\subsection{Surrogate Prediction Performance}
Unless otherwise specified, surrogate prediction results are reported from a single run, while optimization results are averaged over multiple random seeds.

We first evaluate the predictive performance of the QUBO surrogate across different dataset sizes and latent dimensions (see Table \ref{tab:surrogate_spearman}). We observe that increasing dataset size improves surrogate performance, while moderate latent dimensions (e.g., $m=16$ or $m=32$) provide the best balance between expressivity and generalization. Very high latent dimensions (e.g., $m=64$) can lead to overfitting due to the quadratic growth in model parameters.

\begin{table}[t]
\centering
\small
\begin{tabular}{lcccc}
\hline
\textbf{Samples} & \textbf{Dim=8} & \textbf{Dim=16} & \textbf{Dim=32} & \textbf{Dim=64} \\
\hline
1000  & 0.175 & \textbf{0.291} & 0.168 & 0.196 \\
2000  & 0.168 & \textbf{0.257} & 0.235 & 0.225 \\
5000  & 0.236 & 0.287 & \textbf{0.338} & 0.332 \\
10000 & 0.209 & 0.302 & 0.385 & \textbf{0.413} \\
\hline
\end{tabular}
\caption{Test Spearman correlation of the QUBO surrogate across dataset sizes and latent dimensions. Best-performing latent dimension for each dataset size is highlighted in bold.}
\label{tab:surrogate_spearman}
\end{table}

\subsection{Optimization Performance}

We next evaluate the ability of different optimization methods to identify high-fitness regions in the latent space. Results are averaged over multiple random seeds (see Table \ref{tab:optimization}). We find that all optimization methods are able to improve surrogate scores, but their effectiveness in identifying high-fitness sequences varies. In particular, simulated annealing and genetic algorithms achieve strong improvements in the surrogate objective, while random search remains competitive due to the relatively low dimensionality of the latent space. Greedy hill climbing tends to preserve proximity to the training data manifold, which can result in higher true fitness in some cases.

\begin{table*}[t]
\centering
\small
\begin{tabular}{lccc}
\hline
\textbf{Method} & \textbf{Improvement} & \textbf{NN True Fitness} & \textbf{NN Percentile} \\
\hline
Simulated Annealing & \textbf{1.529 $\pm$ 0.239} & 3.675 $\pm$ 0.192 & 84.65 $\pm$ 16.93 \\
Genetic Algorithm   & \textbf{1.529 $\pm$ 0.239} & 3.675 $\pm$ 0.192 & 84.65 $\pm$ 16.93 \\
Random Search       & 1.448 $\pm$ 0.276 & 3.602 $\pm$ 0.233 & 77.45 $\pm$ 19.95 \\
Greedy Hill Climb   & 1.127 $\pm$ 0.510 & \textbf{3.723 $\pm$ 0.084} & \textbf{88.21 $\pm$ 8.68} \\
Latent BO           & -0.104 $\pm$ 0.523 & 3.216 $\pm$ 0.746 & 64.93 $\pm$ 21.54 \\
\hline
\end{tabular}
\caption{Multi-seed optimization results on the GFP benchmark using Q-BioLat with PCA-based binary latent representations (10{,}000 samples, 16 latent bits). Improvement denotes surrogate score gain relative to the starting point. NN True Fitness and NN Percentile are computed from the nearest retrieved training sequence to the optimized latent code. Best values in each column are highlighted in bold.}
\label{tab:optimization}
\end{table*}

\subsection{Effect of Latent Dimension}

\begin{figure*}[t]
\centering

\begin{subfigure}[t]{0.48\linewidth}
    \centering
    \includegraphics[width=\linewidth]{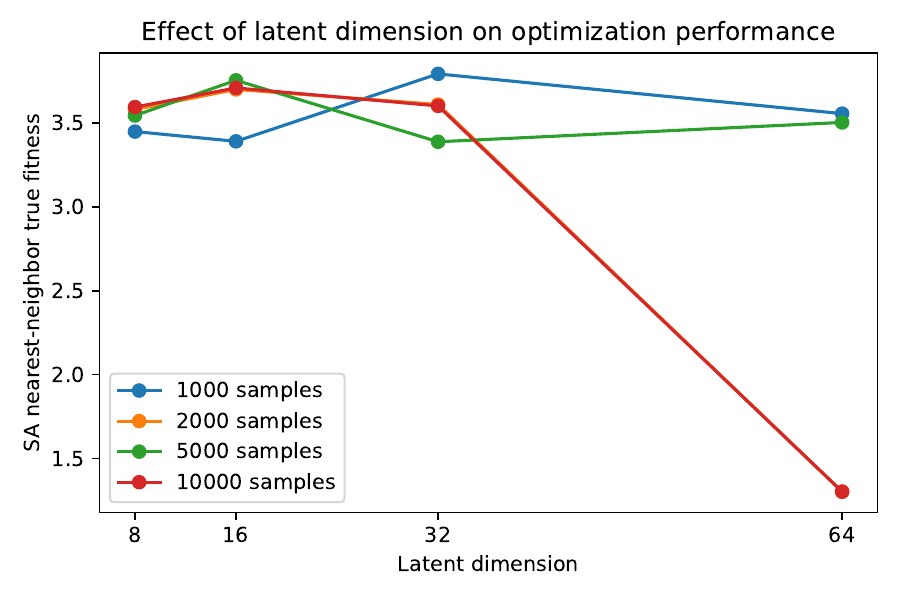}
    \caption{Optimization performance (SA NN true fitness)}
    \label{fig:latent_dim_opt}
\end{subfigure}
\hfill
\begin{subfigure}[t]{0.48\linewidth}
    \centering
    \includegraphics[width=\linewidth]{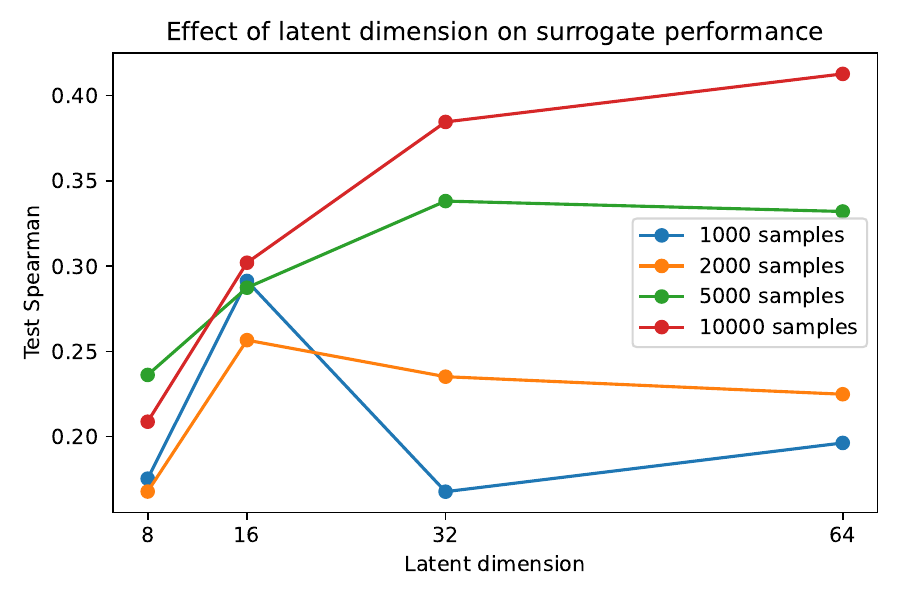}
    \caption{Surrogate performance (test Spearman)}
    \label{fig:latent_dim_spearman}
\end{subfigure}

\caption{Effect of latent dimension on optimization and surrogate performance on the GFP benchmark. Each curve corresponds to a different dataset size. Moderate latent dimensions (e.g., 16–32) provide a favorable trade-off between predictive accuracy and optimization stability.}
\label{fig:latent_dim}
\end{figure*}

To understand the role of representation dimensionality, we analyze optimization performance as a function of latent dimension (see Figure \ref{fig:latent_dim}). We observe that low-dimensional latent spaces are easier to optimize but may limit expressivity, while higher-dimensional spaces provide richer representations but lead to more challenging optimization landscapes. In particular, we find that $m=16$ and $m=32$ offer a good trade-off between representation quality and optimization stability.

\subsection{Representation Comparison}

We compare random projection and PCA-based methods for constructing binary latent representations. The results are summarized in Table~\ref{tab:representation}.

While both methods achieve nearly identical predictive performance in terms of Spearman correlation, we observe a consistent difference in optimization behavior. In particular, PCA-based representations yield improved nearest-neighbor true fitness and higher percentile rankings of retrieved sequences, indicating more stable and effective search in latent space.

This result highlights an important distinction between predictive accuracy and optimization performance. Although random projection and PCA produce similar surrogate models, the geometric structure of the latent space differs substantially between the two representations. PCA preserves dominant directions of variance in the embedding space, which appear to better align with biologically meaningful variations and thus provide smoother fitness landscapes for combinatorial search. In contrast, random projection introduces more isotropic mixing of features, which can degrade the alignment between latent directions and fitness-relevant variations.

These findings suggest that the quality of latent representations should be evaluated not only in terms of predictive metrics such as Spearman correlation, but also in terms of their induced optimization landscapes. In the context of \modelname, representation geometry plays a critical role in determining the effectiveness of combinatorial optimization, even when surrogate prediction performance remains unchanged.

\begin{table*}[t]
\centering
\small
\begin{tabular}{lccc}
\hline
\textbf{Representation} & \textbf{Spearman} & \textbf{NN True Fitness} & \textbf{NN Percentile} \\
\hline
Random Projection & 0.302 & 3.622 $\pm$ 0.123 & 74.76 $\pm$ 13.35 \\
PCA Projection & 0.302 & \textbf{3.675 $\pm$ 0.192} & \textbf{84.65 $\pm$ 16.93} \\
\hline
\end{tabular}
\caption{
Comparison of latent representation methods on the GFP benchmark (10{,}000 samples, 16 latent bits). Although both methods achieve similar Spearman correlation, PCA yields improved optimization performance, highlighting the role of representation geometry in guiding combinatorial search.
}
\label{tab:representation}
\end{table*}

\subsection{Discussion of Optimization Behavior}

Across experiments, we observe that surrogate improvement does not always directly correlate with true fitness gains. This highlights a key challenge in latent space optimization: the surrogate landscape may contain regions that are favorable under the model but less aligned with true fitness. Methods that constrain search near the training manifold, such as greedy hill climbing, can sometimes achieve higher true fitness despite smaller surrogate improvements. Overall, these results demonstrate that Q-BioLat provides a viable framework for exploring protein fitness landscapes in binary latent space, while also revealing important trade-offs between representation quality, model complexity, and optimization strategy.
\section{Conclusion} \label{sec:conclusion}

We introduced \modelname, a framework for modeling and optimizing protein fitness landscapes in binary latent spaces. By transforming protein language model embeddings into compact binary representations and fitting a QUBO surrogate, \modelname enables protein fitness optimization to be formulated as a structured combinatorial problem. This formulation allows the use of classical optimization methods such as simulated annealing and genetic algorithms, while remaining directly compatible with emerging quantum annealing hardware.

Through experiments on the ProteinGym GFP benchmark, we demonstrated that \modelname can identify high-fitness regions of the sequence space, with optimized latent codes retrieving sequences near the top of the observed fitness distribution. Our results further reveal important trade-offs between latent dimensionality, surrogate generalization, and optimization stability, as well as a key distinction between predictive accuracy and optimization effectiveness. In particular, we showed that representation geometry significantly impacts optimization performance, even when surrogate prediction metrics remain similar.

Overall, this work highlights the potential of binary latent representations for bridging protein representation learning and combinatorial optimization. By expressing protein fitness landscapes as QUBO problems, \modelname opens a new direction toward quantum-compatible protein engineering. Future work includes developing learned binary representations, improving surrogate models beyond pairwise interactions, and exploring integration with quantum annealing hardware to enable scalable, quantum-assisted protein design.

\bibliography{tex/main}
\bibliographystyle{unsrtnat}

\end{document}